\pdfoutput=1

\documentclass[11pt]{article}

\usepackage[]{acl}

\usepackage{times}
\usepackage{latexsym}

\usepackage[T1]{fontenc}

\usepackage[utf8]{inputenc}

\usepackage{microtype}

\usepackage{inconsolata}

\usepackage{inconsolata}
\usepackage{graphicx}
\usepackage{colortbl}
\usepackage{bbm}
\usepackage{amsfonts}
\usepackage[normalem]{ulem}
\useunder{\uline}{\ul}{}
\usepackage{amssymb}
\usepackage{amsfonts}
\usepackage{framed}
\usepackage{color}
\usepackage{multicol}
\usepackage{multirow}
\usepackage{makecell}
\usepackage{amsmath}
\usepackage{booktabs}
\usepackage{array}
\usepackage{paralist}
\usepackage{algorithm}
\usepackage{bbm}
\usepackage{dsfont}
\usepackage{adjustbox}
\usepackage{diagbox}
\usepackage{algpseudocode}
\usepackage{multicol}
\usepackage{wrapfig}
\usepackage{enumitem}
\usepackage{caption}
\usepackage[most]{tcolorbox}
\usepackage{svg}
\usepackage{arydshln}
\usepackage{arydshln}
\usepackage{bm}

\title{Multi-Agent Sampling: Scaling Inference Compute for Data Synthesis \\ with Tree Search-Based Agentic Collaboration}

\author{Hai Ye$^{1, 2}$  \ \  \ Mingbao Lin$^2$\thanks{Corresponding Author.} \ \ \ Hwee Tou Ng$^1$ \ \ \ Shuicheng Yan$^2$\\
    $^1$National University of Singapore \ \ \ \  $^2$Skywork AI \\
    \texttt{yehai@comp.nus.edu.sg}\quad \texttt{linmb001@outlook.com} \\
    \texttt{nght@comp.nus.edu.sg} \quad \texttt{shuicheng.yan@kunlun-inc.com}
  \\}

\begin{document}
\maketitle

\begin{abstract}

Scaling laws for inference compute in multi-agent systems remain under-explored compared to single-agent scenarios. This work aims to bridge this gap by investigating the problem of data synthesis through multi-agent sampling, where synthetic responses are generated by sampling from multiple distinct language models. 
Effective model coordination is crucial for successful multi-agent collaboration. Unlike previous approaches that rely on fixed workflows, we treat model coordination as a multi-step decision-making process, optimizing generation structures dynamically for each input question. 
We introduce Tree Search-based Orchestrated Agents~(TOA), where the workflow evolves iteratively during the sequential sampling process. To achieve this, we leverage Monte Carlo Tree Search (MCTS), integrating a reward model to provide real-time feedback and accelerate exploration. 
Our experiments on alignment, machine translation, and mathematical reasoning demonstrate that multi-agent sampling significantly outperforms single-agent sampling as inference compute scales. TOA is the most compute-efficient approach, achieving SOTA performance on WMT and a 72.2\% LC win rate on AlpacaEval. Moreover, fine-tuning with our synthesized alignment data surpasses strong preference learning methods on challenging benchmarks such as Arena-Hard and AlpacaEval\footnote{\url{https://github.com/oceanypt/TOA}}~\footnote{Work done during Hai Ye was an intern at Skywork AI.}.

\end{abstract}

\section{Introduction}

Beyond scaling parameters and data during pre-training of large language models~\cite{DBLP:journals/corr/abs-2001-08361}, recent research has increasingly focused on scaling compute at inference time~\cite{DBLP:journals/corr/abs-2407-21787,DBLP:journals/corr/abs-2408-03314}. Unlike rapid problem solving, scaling inference compute generates more output tokens, allowing the model to address problems more deliberately, potentially engaging in planning, reasoning, or refinement before finalizing an answer. Such an emerging scaling paradigm in inference significantly enhances the model's problem solving and data synthesis abilities.

\begin{figure}[t]
\centering
\includegraphics[width=0.9\columnwidth]{ 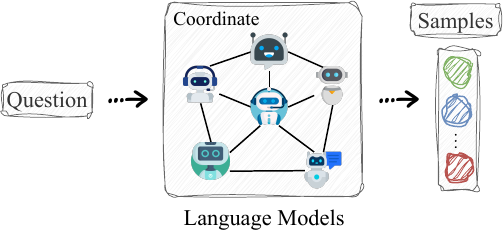} 
\vspace{-0.5em}
\caption{Illustration of multi-agent sampling: We scale the number of samples (compute) generated from a group of distinct language models for each question.}
\vspace{-1.0em}
\label{fig:setting_mas}
\end{figure}

Scaling laws of inference compute have been widely studied in the single-agent scenarios~\cite{DBLP:journals/corr/abs-2407-21787,bansal2024smaller}. However, scaling compute from multi-agent systems remains under-explored. We hypothesize that a multi-agent system offers a higher performance ceiling compared to a single-agent system. This is based on the observation that different language models exhibit varying strengths, and leveraging the unique advantages of each model can lead to enhanced overall capabilities. To address this gap, we propose to study multi-agent sampling that scales the number of generated samples (compute) from multiple distinct language models~(see Figure\,\ref{fig:setting_mas}).

To scale inference compute, we adopt the widely used best-of-$N$ sampling method, which generates $N$ outputs for each input prompt~\cite{DBLP:conf/iclr/0002ZJKSLL24,DBLP:journals/corr/abs-2308-08998}. In multi-agent-based sampling, the primary challenge is effectively coordinating models to achieve compute-optimal generation. We begin by revisiting and formalizing previous methods for model fusion, such as mixture of agents~\cite{DBLP:journals/corr/abs-2406-04692}, within a unified framework. These methods typically rely on fixed workflows with a modular approach to model combination, where the responses of other models are encoded as input context to generate new outputs. However, fixed structures fail to account for question-specific variability, as the optimal structure varies across different questions. To overcome this limitation, we propose Tree Search-based Orchestrated Agents~(TOA), which treat model coordination as a multi-step decision-making process, aiming to maximize the rewards of sequentially generated responses. TOA utilizes Monte Carlo Tree Search (MCTS)~\cite{DBLP:journals/tciaig/BrownePWLCRTPSC12} to dynamically optimize generation workflows for each input question. This process is enhanced by a reward model that provides real-time feedback, enabling more efficient exploration and adaptation.

In our experiments, we study alignment, machine translation, and mathematical reasoning with varying number of parameters. By scaling sample generation and recording FLOPs as a compute metric, we find that multi-agent compute scaling is more efficient than single-agent ones. Our method, TOA, achieves a 72.2\% LC win rate on AlpacaEval 
and sets new SOTA performance on WMT test sets. Additionally, fine-tuning with our synthetic alignment data outperforms strong baselines like SimPO~\cite{DBLP:journals/corr/abs-2405-14734} on AlpacaEval and Arena-Hard\footnote{\url{https://github.com/lmarena/arena-hard-auto}}. Our contributions are as follows:
\begin{itemize}[itemsep=-5pt, topsep=0pt, leftmargin=0.3cm]
    \item We study the under-explored problem of scaling inference compute with multi-agent systems.
    \item We propose a novel method TOA to coordinate multiple language models effectively.
    \item We conduct extensive experiments to reveal scaling laws of multi-agent systems and demonstrate the compute efficiency of our approach.
\end{itemize}

\section{Related Work}

\noindent{\textbf{Scaling Inference Compute.}} \ 
Recent studies explore the trade-off between compute and performance in scaling test-time inference. \citet{DBLP:journals/corr/abs-2407-21787} use repeated random sampling to enhance performance of large language models, relying on effective verifiers to identify correct answers. \citet{DBLP:journals/corr/abs-2408-03314} show that scaling inference computations is more effective than scaling pre-training. Similarly, \citet{bansal2024smaller} find that smaller models with increased inference computation generate better synthetic data than larger models. \citet{wu2024empirical} examine inference scaling laws in mathematical problem solving. Unlike these single-agent studies, our work uniquely investigates scaling laws in multi-agent systems.

\noindent{\textbf{Multi-Agent Learning.}} Combining diverse agents for problem solving is challenging. Some studies assign distinct roles to language models to facilitate collaboration~\cite{DBLP:conf/iclr/ChenSZ0YCYLHQQC24}. LLM debating and fusion are widely adopted, where the output of one model is passed to another for further refinement~\cite{DBLP:conf/nips/LiHIKG23,DBLP:conf/emnlp/Liang0JW00Y0T24,DBLP:conf/emnlp/XiongD00023,DBLP:conf/acl/ChenSB24}. \citet{DBLP:conf/acl/Jiang0L23} train models to fuse the outputs of multiple models, while \citet{DBLP:conf/icml/Du00TM24} enable models to engage in debates with each other. \citet{DBLP:journals/corr/abs-2406-04692} propose a fully connected network to fuse and refine model outputs. Optimizing generation structures is critical to reducing compute and enhancing performance. \citet{DBLP:conf/naacl/LuYLLYZZ24} train routing models to predict the best model for a given input question. Similarly, \citet{liu2023dynamic} perform task-level structure optimization, which requires retraining the structure for every new task. In contrast, we focus on instance-level structure optimization using MCTS, which enables efficient and dynamic tree buildup for any input question. More related work on synthetic data generation and MCTS is given in Appendix~\ref{sec:more_related_work}.

\begin{figure*}[t]
\centering
\begin{minipage}{\textwidth}
    \centering
        \includegraphics[width=0.95\linewidth]{ 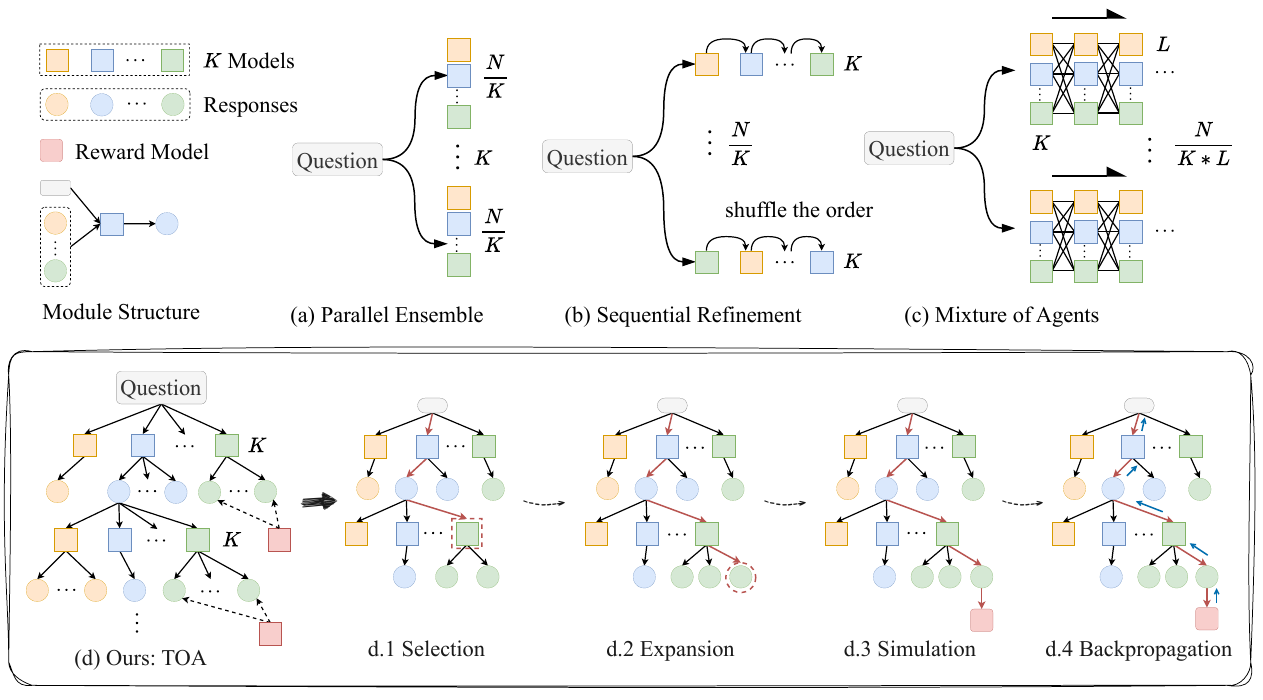}
\end{minipage}
\caption{Illustration of previous methods~(\texttt{a}, \texttt{b} and \texttt{c}) and our method TOA for multi-agent sampling. The methods used to sample $N$ responses per question share the same model structure but differ in coordination strategies. TOA casts the problem as multi-step decision making and uses Monte Carlo Tree Search~(MCTS) to decide which model and response to use to generate a new sample. Stages d.$1$ to d.$4$ display how a new sample is generated.}
\label{fig:method}
\end{figure*}

\section{Preliminaries}
\subsection{Multi-Agent Sampling}
As shown in Figure\,\ref{fig:setting_mas}, we study multi-agent sampling which scales inference compute from multiple distinct language models. It uses best-of-$N$ sampling, where for each question, it generates multiple output samples. Formally, we have $K$ models for generation with different model parameters, each denoted by $\pi_k$. Given an input prompt $\bm{x}$, we sample $N$ responses $\{\bm{y}_1, \cdots, \bm{y}_N \}$ from the $K$ models. We aim to achieve compute-optimal sampling to maximize the sample rewards given a compute budget.

\subsection{Module Structure for Combination}
To combine multiple language models, we let the models \emph{interact} with each other, that is, a language model will utilize the outputs of other models to generate new responses. This method establishes a modular structure capable of supporting diverse generative frameworks~(see Figure\,\ref{fig:method}).

Formally, to obtain $N$ output samples in total, when generating the $i$-th response, 
we choose a model $k$ from the pool and prepare the input context with the output responses of other models:
\begin{equation}
    \bm{y}_i \sim \pi_k(\cdot | \bm{x}, \bm{z}_i),
    \label{eq:unit_design}
\end{equation}
where $\bm{z}_i$ is a concatenation of certain outputs from the past $i-1$ generated samples. This process is similar to iterative inference or refinement, which is expected to improve sample quality~\cite{DBLP:conf/nips/MadaanTGHGW0DPY23}. 
Since the context $\bm{z}_i$ contains responses from other models, the capabilities of other models can be combined and fused.

Depending on the specific design, $\bm{z}_i$ may encompass responses from one to $K$ models out of all the $K$ models, or it could potentially be devoid of any responses. 
In $\S$~\ref{sec:methods}, we revisit previous methods for combining language models, whose unit structure can be formalized using Eq.\,(\ref{eq:unit_design}).

\section{Prior Methods}\label{sec:methods}
In this section, we revisit previous methods for model combination (see Figure\,\ref{fig:method}).

\subsection{Parallel Ensemble}
For each input, the simplest approach to sampling \(N\) responses from \(K\) models is to draw an equal number of responses from each model in parallel. Specifically, for each model \(k\), we sample \(\frac{N}{K}\) responses using the formula \(\mathbf{y} \sim \pi_k(\cdot | \mathbf{x})\). Temperature sampling is employed in this process to generate diverse output samples.

\subsection{Sequential Refinement}
In this setup, to obtain the $i$-th response, the input context $\bm{z}_i$ only contains one already generated response from one specific model. Sequential Refinement studied in \cite{DBLP:conf/nips/MadaanTGHGW0DPY23,DBLP:journals/corr/abs-2408-03314} aligns well with this setup, where the model iteratively modifies the last generated response. 
In the multi-agent setting, a current model can refine the output of a different preceding model. Since we have $K$ models, one complete sequential refinement will yield $K$ samples. To generate $N$ responses, we can perform $\frac{N}{K}$ parallel generation operations, with each thread executing the sequential refinement. To enhance diversity, each thread will randomize the model order.

\subsection{Mixture of Agents}
Model debating and fusion have been widely studied to coordinate multiple agents~\cite{DBLP:conf/acl/Jiang0L23,DBLP:conf/icml/Du00TM24}, which train a fusion model to fuse responses of multiple models or stack multiple layers for iterative fusion and refinement. This line of work aims for one model to encode the outputs from all the models to generate new samples, where the input context $\mathbf{z}$ comprises $K$ responses, with each response from a different model.

A representative work is Mixture-of-Agents (MoA)~\cite{DBLP:journals/corr/abs-2406-04692}, which adopts a structure akin to a fully connected network, where each node represents an individual model. It consists of multiple layers and each layer contains $K$ models, generating $K$ outputs that feed into the subsequent layer. Within each layer, each model refines the outputs of the $K$ models from the preceding layer, facilitating a progressive refinement. Suppose there are $L$ layers in MoA, we can obtain $L * K$ samples by going through the full network for one pass. We repeat the iterations of flowing over the network until we obtain $N$ samples. 

{
The methods described above rely on fixed structures, either linear or graph-based, to integrate different models. These approaches utilize fixed structures for all input scenarios. It overlooks the variation in questions.  Varying input prompts might benefit from different, potentially more optimal structure configurations. 
}

\section{Tree Search-Based Orchestrated Agents}
We present Tree Search-based Orchestrated Agents (TOA), capable of optimizing the generation workflow for each input question. 

We treat model coordination as a multi-step decision-making process, which generates a sequence of samples given the input question $\bm{x}$. It sequentially selects a model $k$ from $K$ predefined models, and a response $\bm{y}_j$ from past generations. We formalize the process as a Markov decision process with ($S$, $A$, $T$, $R$) intending to maximize the cumulative reward:
\begin{equation}
    J(\pi) =\mathbb{E}\Big[ \sum_{i=1}^N R(s_i, a_i) \Big],
\end{equation}
and we define the Markov decision process below: 

\noindent{$\bullet$ \textbf{State Space ($S$):}} \ The state in step $i$ is $s_i = (\bm{x}, Y)$, where $\bm{x}$ is the fixed input question and $Y = \{\bm{y}_1, \cdots, \bm{y}_{i-1}\}$ is a set of samples generated in the first $i-1$ steps.

\noindent{$\bullet$ \textbf{Action Space ($A$):}} \ In each step $i$, an action $a_i = ({k}, \bm{y}_j)$ consists of selecting a model $k \in \{1, \cdots, K\}$ and a previously generated response $\bm{y}_j \in Y$, where $1 \leq j \leq i-1$. For $i=1$, no prior responses exist, so $\bm{y}_j$ is empty.

\noindent{$\bullet$ \textbf{Transition Function ($T$):}} \ From state $s_i$ to state $s_{i+1}$, the selected model $k$ generates a new response $\bm{y}_i = \pi_k(\bm{x}, \bm{y}_j)$. Then this new response $\bm{y}_i$ is added to $Y$.

\noindent{$\bullet$ \textbf{Reward Function ($R$):}} \ The reward function evaluates the quality of each generated response as $R(s_i, a_i) = R(\bm{x}, \bm{y}_i)$, where $\bm{y}_i$ is the newly generated response in step $i$.

\subsection{Reward-Guided MCTS}
As shown in Figure\,\ref{fig:method}, we use MCTS as our decision-making framework, where the tree starts at the root node and proceeds to alternate between model and response layers. The width of the model layer corresponds to the number of models $K$. Conversely, the width of the response layer $w$ is contingent upon the configuration at hand. Generating the $i$-th response is a multi-stage process, comprising selection, expansion, simulation, and backpropagation.

\noindent{\textbf{(a) Selection.}} \ In generating the $i$-th response, we are tasked with determining the action $a_i = (k, \bm{y}_j)$: the model $k$ for generation and the response $\bm{y}_j$ from past generations. We traverse the tree to select the optimal model node associated with the model $k$, and the response $\bm{y}_j$ present in its \emph{parent} node. The search process adapts operations to each node type encountered:
\begin{itemize}[itemsep=-5pt, topsep=0pt, leftmargin=0.3cm]
    \item \textbf{Model Nodes:} Upon reaching a model node, we selectively prune its child nodes (responses), retaining only those with the highest reward scores.
    
    \item \textbf{Response Nodes:} In contrast, when encountering a response node, we refrain from pruning its child nodes (models), allowing for a broader exploration of subsequent model options.
\end{itemize}

In the search tree, starting from the root, we keep choosing child nodes until we find one that has not yet been fully explored (i.e., not all possible actions have been tried). 
We use the Upper Confidence Bound (UCB) strategy to pick a child node, balancing exploration and exploitation. The UCB score is computed as:
\begin{equation}
\text{UCB} = \frac{v}{n} + \alpha \cdot \sqrt{\frac{2 \ln N}{n}},
\end{equation}
where $v$ is the cumulative reward for a node, $n$ is the visit count of the node, and $\alpha$ is a hyper-parameter that controls exploration. $N$ is the maximum number of responses. 

When a response node expands to include $K$ model nodes, which constitute a predetermined set, it is considered fully expanded. Upon completion of the selection process, we identify a model node designated for the generation task. Simultaneously, its parent response will be selected for refinement.

\begin{figure*}[t]
\centering
\begin{minipage}{0.45\linewidth}
    \centering
        \includegraphics[width=\linewidth]{ 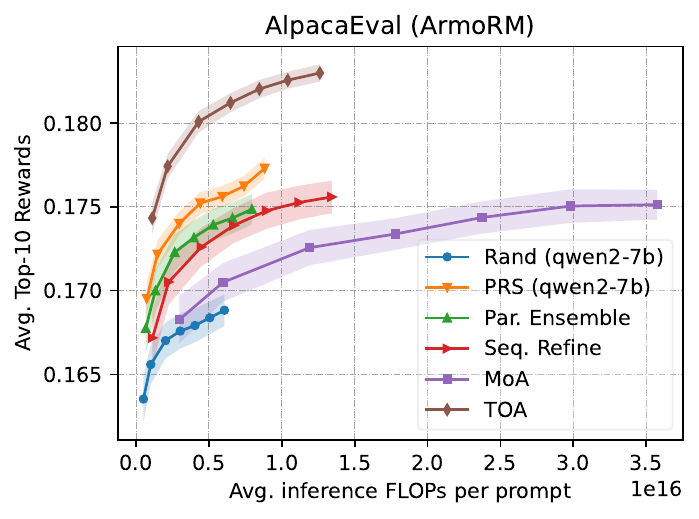}
\end{minipage}
\begin{minipage}{0.45\linewidth}
    \centering
        \includegraphics[width=\linewidth]{ 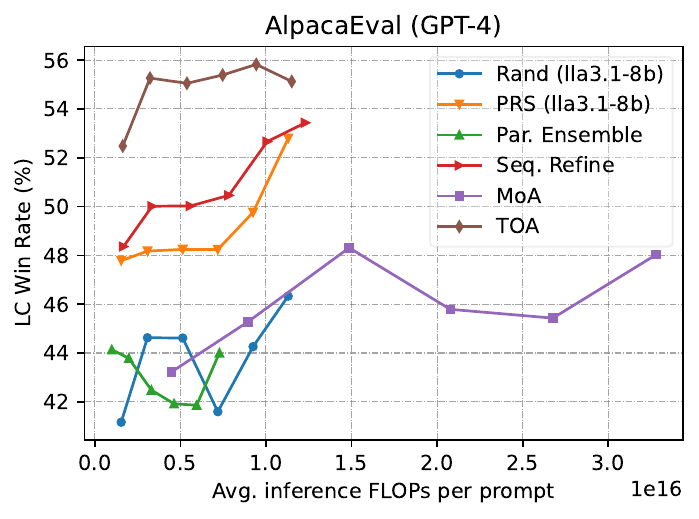}
\end{minipage}
\caption{Scaling results for alignment on AlpacaEval with 4 small-scale models, varying the number of samples per prompt, i.e., 64, 128, 256, 384, 512, 640, and 768. \textbf{Left:} ArmoRM is used as the reward model for guidance and evaluation. \textbf{Right:} The best response is selected for GPT-4 evaluation to calculate the length-controlled win rate. We only present the best models for Rand and \emph{PRS}. 
AlpacaEval is randomly down-sampled to 200 prompts.}
\label{fig:scaling_alignment}
\end{figure*}

\noindent{\textbf{(b) Expansion.}} \ 
Next, we expand the response and model nodes with distinct operations:
\begin{itemize}[itemsep=-5pt, topsep=0pt, leftmargin=0.3cm]
    \item \textbf{Response Node:} From the response node, we directly expand to new nodes, one for each of the $K$ models. Moreover, model nodes that are immediate children of the root are merged into the response node by inheriting their visit counts and reward values. This method allows us to conduct operations without additional refinement, important in scenarios where further refinement cannot bring improvement.

    \item \textbf{Model Node:} After a model node is selected, we expand this model node by generating a new response using:
    \begin{equation}
        \bm{y}_i \sim \pi_k(\cdot |\bm{x}, \bm{y}_j),
    \end{equation}
    where $\bm{y}_j$ is the response from the parent response node. However, if the parent does not contain a response---in particular when the parent is a root node or the current model node is copied from the children of the root---then $\bm{y}_j$ will be empty. 
\end{itemize}

\noindent{\textbf{(c) Simulation.}} \ Then we determine the reward $r_i$ for newly generated response by leveraging a reward model $R$:
\begin{equation}
    r_i \gets R(\bm{x}, \bm{y}_i),
\end{equation}
which can provide real-time feedback and speed up exploration.

\noindent{\textbf{(d) Backpropagation.}} \ Lastly, an update occurs along the path that extends to the leaf node, updating the visit counts and the cumulative rewards with the following adjustments:
\begin{align}
    v &\gets v + r_i, \\
    n &\gets n + 1.
\end{align}

The above stages of a$-$d repeat until we have successfully collected $N$ responses, present within all the response nodes in our decision tree.

\begin{figure*}[t]
\centering
\begin{minipage}{0.45\linewidth}
    \centering
        \includegraphics[width=\linewidth]{ 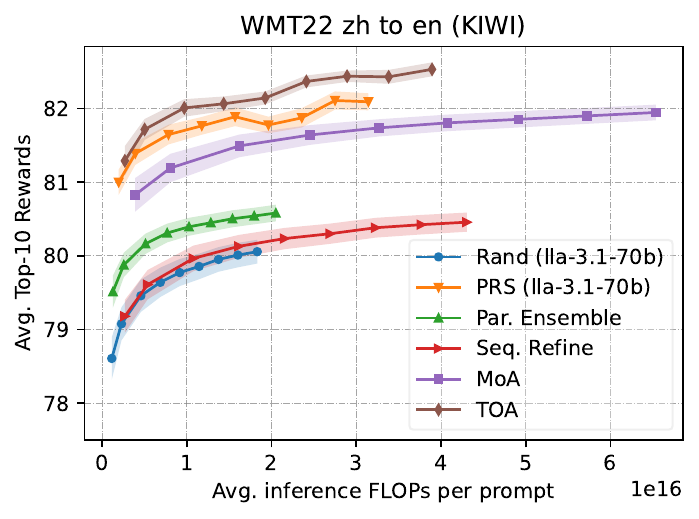}
\end{minipage}
\begin{minipage}{0.45\linewidth}
    \centering
        \includegraphics[width=\linewidth]{ 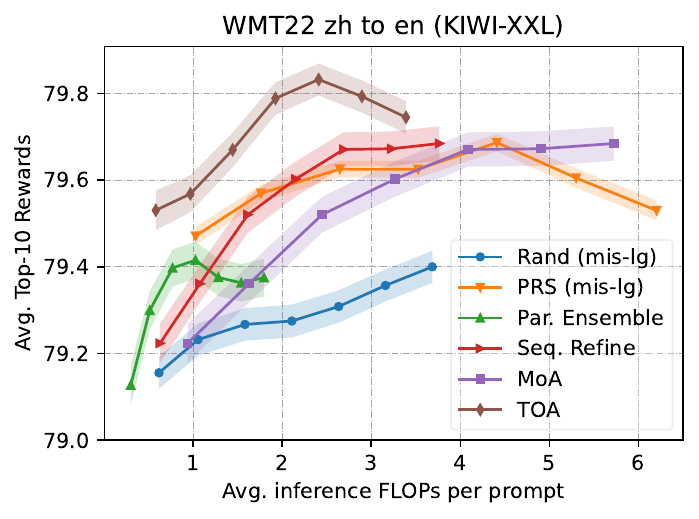}
\end{minipage}
\caption{
Scaling results for machine translation~(zh to en) on WMT'22 using 5 large-scale models. The number of translations varies among 80, 160, 320, 480, 640, 800, 960, 1120, and 1280. \textbf{Left:} KIWI is used as the reward model for guidance and evaluation. \textbf{Right:} The best translation is selected and re-evaluated by KIWI-XXL~(\texttt{Unbabel/wmt23-cometkiwi-da-xxl}), where moving average is applied to smooth the curves. Only the best models for Rand and \emph{PRS} are presented. The test set is randomly down-sampled to 200 test samples.
}
\label{fig:scaling_nmt}
\end{figure*}

\begin{figure*}[t]
\centering
\begin{minipage}{0.45\linewidth}
    \centering
        \includegraphics[width=\linewidth]{ 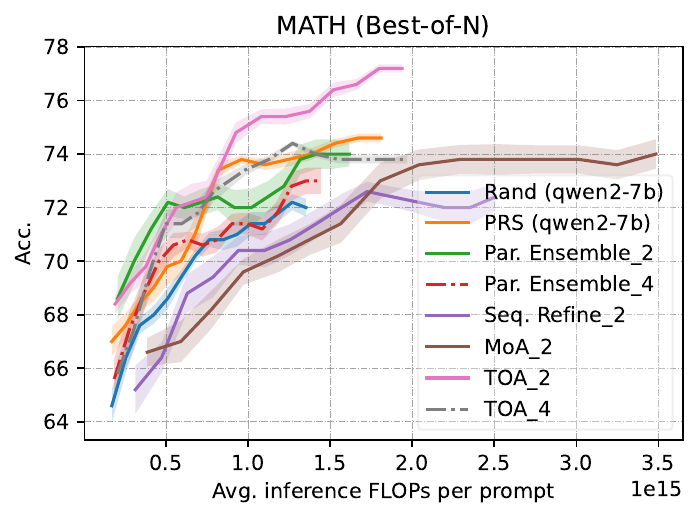}
\end{minipage}
\begin{minipage}{0.45\linewidth}
    \centering
        \includegraphics[width=\linewidth]{ 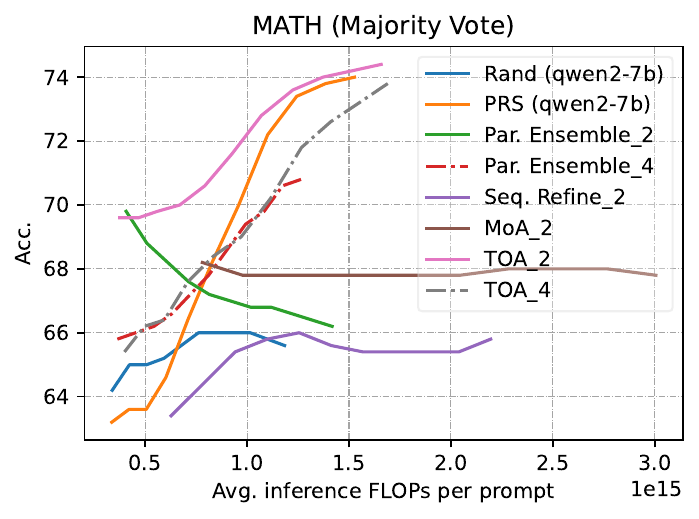}
\end{minipage}
\caption{
Scaling results for math where the 4 small-scale models are combined. We generate solutions iteratively until we obtain 128 samples, then analyze the results of the first $N$ solutions. \textbf{Left:} A single solution is selected from the first $N$ solutions, and we calculate the average accuracy of the top 5 solutions with the highest rewards to reduce the variance. \textbf{Right:} The final answer is determined using majority voting among the first $N$ solutions and moving average is applied to smooth the curves. The number next to ``$\_$'' indicates the number of models combined; for 2, we chose Llama-3.1-8B-Inst and Qwen2-7B-Inst. 
}
\label{fig:scaling_math}
\end{figure*}

\section{Experiments}
\subsection{Setup}

\noindent{\textbf{Language Models.}} We examine two groups of models, with small-scale and large-scale model parameters. Within each group, we expect comparable performance among the models. 
In the small-scale group, we combine four models: Llama-3.1-8B-Instruct, Qwen2-7B-Instruct, Mistral-7B-Instruct-v0.2, and Yi-1.5-9B-Chat-16K. 
In the large-scale group, we combine five models: Llama-3.1-70B-Instruct~(Lla), Mistral-Large-Instruct-2407~(Mis), Qwen2-72B-Instruct~(Qwen), Mixtral-8x22B-Instruct-v0.1~(Mix), and Wizardlm-2-8x22B~(Wiz). Appendix~\ref{sec:app_lang_model} provides their HuggingFace links.

\noindent{\textbf{Datasets.}} \ We study three practical tasks: alignment, machine translation~(MT), and math reasoning. We evaluate alignment on AlpacaEval~\cite{li2023alpacaeval} and Arena-Hard~\cite{DBLP:journals/corr/abs-2406-11939} datasets. For MT, WMT'21 and WMT'22 are used. We test math reasoning on MATH~\cite{DBLP:conf/nips/HendrycksBKABTS21}. Finally, we synthesize alignment data with the prompts from Ultrafeedback~\cite{DBLP:conf/icml/CuiY0YH0NXXL0024}. Appendix~\ref{statistics} provides more statistics.

\noindent{\textbf{Reward Models.}} We utilize a reward model to guide the search process. For alignment, we employ ArmoRM~\cite{DBLP:conf/emnlp/00030X0024} (\texttt{RLHFlow/ArmoRM-Llama3-8B-v0.1}), as it has demonstrated superior performance on the RewardBench benchmark~\cite{DBLP:journals/corr/abs-2403-13787}. In the MT domain, we rely on KIWI~\cite{DBLP:conf/wmt/ReiTGZFMSGACLM22} (\texttt{Unbabel/wmt22-cometkiwi-da}), which is specifically designed for reference-free evaluation. For math reasoning, we adopt Qwen2.5-Math-RM-72B~\cite{DBLP:journals/corr/abs-2409-12122} as the reward model, leveraging its advanced capabilities for this domain.

\noindent{\textbf{Baselines.}} For single-agent sampling, we use random sampling~(Rand) and \emph{PRS}~\cite{DBLP:conf/emnlp/YeN24}. \emph{PRS} employs a tree structure and iterative refinement for data sampling, which has proven to be superior to random sampling. For multi-agent sampling, we compare the baselines discussed in $\S$~\ref{sec:methods}. Note that MoA exemplifies the category of methods referred to as model debating and fusion~\cite{DBLP:conf/icml/Du00TM24}.

\noindent{\textbf{Inference Settings.}} \ We vary the number of samples generated for each question. We calculate FLOPs to measure the cost and record the model performance. We use the method from \citet{hoffmann2022training} to calculate FLOPs. 
Appendix~\ref{sec:setting_inference} provides more details about the settings, and Appendix~\ref{sec:setting_prompt_refine} provides the prompts used.

\begin{table}[!t]
\centering
\Large
\resizebox{\linewidth}{!}{%
\begin{tabular}{lccc}
\toprule
\textbf{Model Name}            & \textbf{LC WR} & \textbf{WR}   & \textbf{|token|} \\ \midrule
Qwen2 72B Instruct             & 38.1           & 29.9          & -                \\
Llama 3.1 70B Instruct         & 38.1           & 39.1          & -                \\
Llama-3.1-405B-Instruct        & 39.3           & 39.1          & -                \\
GPT-4 Turbo (04/09)            & 55.0           & 46.1          & -                \\
GPT-4 Omni (05/13)             & 57.5           & 51.3          & -                \\
Together MoA-Lite              & 59.1           & 56.6          & -                \\
Together MoA                   & 65.4           & 59.9          & -                \\
OpenPipe MoA GPT-4 Turbo       & 68.4           & 63.2          & -                \\ \midrule
\multicolumn{4}{l}{\underline{\textit{{Best-of-160 Sampling}}}}                                  \\
Rand (Llama-3.1-70B-Inst)      & 47.8           & 47.1          & 2004             \\
Rand (Mistral-Large-Inst-2407) & 62.5           & 55.1          & 1809             \\
\emph{PRS} (Llama-3.1-70B-Inst)       & 52.8           & 53.7          & 2094             \\
\emph{PRS} (Mistral-Large-Inst-2407)  & 67.9           & 59.5          & 1787             \\
Par. Ensemble                  & 58.6           & 59.1          & 2092             \\
Seq. Refine                    & 70.0           & 65.6          & 1943             \\
MoA                            & 64.1           & \textbf{73.0} & 2550             \\
TOA~(Ours)                          & \textbf{72.2}  & 68.7          & 1999   \\ \bottomrule
\end{tabular}%
}
\caption{
AlpacaEval 2.0 results show length-controlled win rate (\%) and win rate (\%) with GPT-4-Preview-1106 as both baseline and annotator. Using 5 large-scale models, 160 outputs per prompt are sampled, with ArmoRM-Llama3-8B-v0.1 selecting the best for evaluation.}
\label{tab:alpaca_eval}
\end{table}

\subsection{Results}\label{results}
We analyze compute efficiency across three tasks, explore best-of-\(N\) sampling for AlpacaEval and WMT datasets, and generate synthetic data to fine-tune models for alignment. 
We observe: 

\noindent{\textbf{Multi-agent sampling generally outperforms sampling from a single model.}} Based on the scaling results in Figures\,\ref{fig:scaling_alignment}, \ref{fig:scaling_nmt}, and \ref{fig:scaling_math}, the parallel ensemble method generally demonstrates superior performance compared to random sampling from the best single model. This trend is observed across various tasks and both sets of model combination.

\noindent{\textbf{However, an effective strategy is essential to combine agents.}} 
Without a robust strategy, multi-agent sampling can underperform compared to advanced single-agent methods like \emph{PRS}, which highlights the complexity of model coordination.

\noindent{\textbf{Sequential refinement and mixture-of-agents possess unique strengths.}}
Seq. refinement shows superior compute efficiency over \emph{PRS} for alignment on AlpacaEval, achieving a 70.0 LC win rate with best-of-160 sampling (Table\,\ref{tab:alpaca_eval}) and outperforming strong baselines. For machine translation, MoA surpasses GPT-4 in the reference-free evaluation metric. However, MoA is computationally expensive, as it aggregates the responses of multiple models to generate new output samples, increasing the cost of processing the input context.

\begin{table}[!t]
\centering
\Huge
\resizebox{\linewidth}{!}{%
\begin{tabular}{lccccccc}
\toprule
\textbf{KIWI-XXL}  & \texttt{de}    & \texttt{cs}    & \texttt{is}   & \texttt{zh}    & \texttt{ru}    & \texttt{Avg.}           & \LARGE{FLOPs}        \\
\midrule
Gold Reference     & 78.56          & 83.11          & 85.04         & 74.19          & 79.59          & 80.10          & -                    \\
WMT Winners        & 83.59          & 82.53          & 85.60          & 73.28          & 80.97          & 81.19          & -                    \\
GPT-4              & 84.58          & 83.55          & \textbf{85.90} & 77.65          & 81.34          & 82.60          & -                    \\
CPO                & 83.97          & 83.75          & 85.73         & 77.17          & 81.54          & 82.43          & - \\ \midrule
\multicolumn{7}{l}{\underline{\textit{{Best-of-160 Sampling}}}}                                                                       & \multicolumn{1}{l}{} \\
Rand (lla-3.1-70b) & 85.41          & 85.09          & 84.45         & 79.07          & 82.21          & 83.24          & 3.5                  \\
Rand (Mis-lg) & 85.60          & 84.54          & 83.35         & 79.50          & 82.64          & 83.13          & 7.8                  \\
\emph{PRS} (lla-3.1-70b)  & 85.83          & 85.71         & 84.76         & 79.51          & 82.67          & 83.70          & 6.0                  \\
\emph{PRS} (Mis-lg)  & 86.18          & 85.72          & 84.01         & 79.82          & 82.68          & 83.68          & 13                   \\
Par. Ensemble      & 85.82          & 85.35          & 83.58         & 79.47          & 82.53          & 83.35          & 3.9                  \\
Seq. Refine        & 86.14          & 85.81          & 83.86         & 79.39          & 82.68          & 83.58          & 8.2                  \\
MoA                & 86.20          & 86.13          & 84.87         & 79.87          & 82.80          & 83.97          & 12                   \\
TOA~(Ours)               & \textbf{86.25} & \textbf{86.31} & 84.81         & \textbf{79.89} & \textbf{83.00} & \textbf{84.05} & 7.5       \\ \bottomrule
\end{tabular}%
}
\caption{Results on WMT'21 and '22 with large-scale models. We use KIWI-XXL to evaluate the results. The results in the upper section are from \citet{xu2024contrastive}. Total FLOPs~($\times \text{10}^{18}$) for the entire eval set is presented. Results with reference-based evaluation are in Table\,\ref{table:full_nmt}
}
\label{table:part_nmt}
\end{table}

\noindent{\textbf{TOA is the most compute-optimal.}} Based on the scaling results in Figures\,\ref{fig:scaling_alignment}, ~\ref{fig:scaling_nmt}, and ~\ref{fig:scaling_math}, TOA emerges as the most compute-optimal method across all three tasks and both sets of models. To analyze the scaling behavior of TOA, we fit its scaling curve in Figure\,\ref{fig:scaling_alignment} (left) using a function where the input is FLOPs ($C$) and the output is the avg. top-10 reward ($R$). The function takes the form \( R = a \cdot \log_{10}(C)^2 + b \cdot \log_{10}(C) + c \). After fitting, the parameters are determined to be \( a = -0.0031 \), \( b = 0.11 \), and \( c = -0.71 \). This learned function provides a reliable means to predict the reward for a given compute budget. 

In terms of the MATH eval set, as presented in Figure\,\ref{fig:scaling_math}, we observe that combining all four small-scale models does not outperform combining the top two models (i.e., Qwen2-7B-Inst and Llama-3.1-8B-Inst) (also see Figure\,\ref{fig:math_result_differ_model}), which suggests that successful multi-agent sampling requires the combined models to achieve relatively consistent performance.

\noindent{\textbf{Dynamic model coordination enables TOA.}} In Figure\,\ref{fig:refine_path_alpaca_eval} and Figure\,\ref{fig:refine_path_mt} in the Appendix, we analyze the refinement paths with the highest rewards, examining their categories and proportions, as well as the frequency with which specific models follow one another along these optimal paths. Our findings reveal that the top 20 optimal paths represent only a small fraction of the total, highlighting the diversity of effective refinement strategies. 
Notably, the likelihood of a successor model being identical to its predecessor in optimal paths is quite low. 
This suggests that using diverse models for refinement typically yields better results.

\begin{figure}[!t]
\centering
\begin{minipage}{\linewidth}
    \centering
        \includegraphics[width=\linewidth]{ 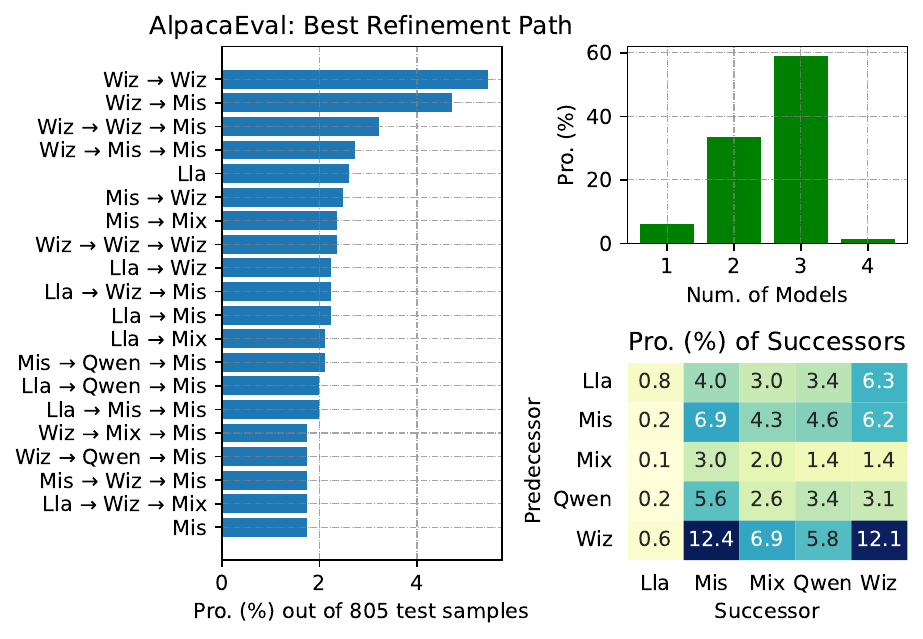}
\end{minipage}
\caption{For TOA with 5 large models, we identify the best refinement paths per prompt that maximize reward. Among the best paths, we present the top 20 most frequent paths, the number of models used for generation, and the successor proportions for each predecessor.
}
\label{fig:refine_path_alpaca_eval}
\end{figure}

\begin{figure}[!t]
\centering
\begin{minipage}{\linewidth}
    \centering
        \includegraphics[width=\linewidth]{ 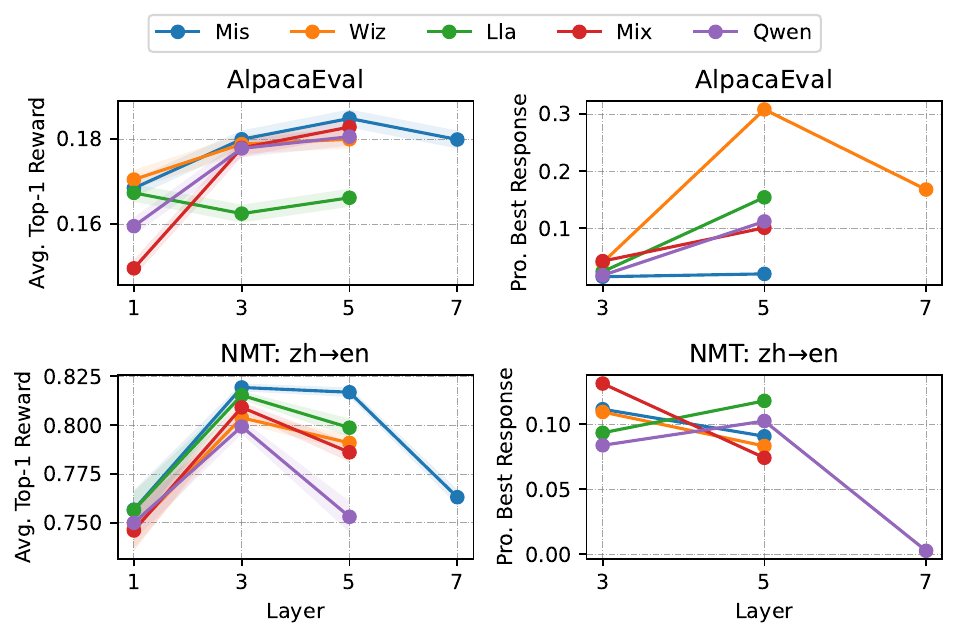}
\end{minipage}
\caption{With the large-scale models, we report the average highest reward on each layer of each model (\textbf{left}), and the proportion of the best response over all samples existing in the layers of each model (\textbf{right}).}
\label{fig:reward_layer_by_layer}
\end{figure}

\begin{table}[t]
\centering
\Huge
\resizebox{\linewidth}{!}{%
\begin{tabular}{lcccc}
\toprule
                             & \multicolumn{2}{c}{\textbf{AlpacaEval}}     & \textbf{Arena-Hard}  & \textbf{FLOPs} \\
                             & \textbf{LC WR}       & \textbf{WR}          & \textbf{WR}          &      $\times 10^{15}$                           \\ \midrule
\multicolumn{5}{l}{\textit{\underline{Llama-3-8B-Inst + Syn. Data}}}  \\
DPO$^\dagger$                  & 48.2                 & 47.5                 & 35.2                 & -                               \\
SimPO$^\dagger$                        & 53.7                 & 47.5                 & 36.5                 & -                               \\ \midrule
Yi-1.5-9B-Chat-16K           & 23.3 & 25.8 & 28.4 & -                               \\
Mistral-7B-Instruct-v0.2     & 17.1                 & 14.7                 & 12.6                 & -                               \\
Qwen2-7B-Instruct            & 21.2                 & 19.5                 & 23.6                 & -                               \\
Llama-3-8B-Instruct        & 24.8                 & 23.6                 & 19.7                 & -                               \\ \midrule
\multicolumn{5}{l}{\textit{\underline{Llama-3-8B-Inst + Syn. Data + SFT}}}                                                                            \\
Rand (Lla-3.1-8B-Inst) & 25.3                 & 23.8                 & 21.7                 & 2.0                        \\
\emph{PRS} (Lla-3.1-8B-Inst)  & 30.4                 & 30.5                 & 24.4                 & 2.9                         \\
Par. Ensemble                & 28.6                 & 26.9                 & 24.9                 & 2.0                         \\
Seq. Refine                  & 34.0 & 30.4 &  21.9  & 3.3            \\
MoA                          & 29.8                 & 32.1                 & 24.4                 & 7.5                         \\
TOA~(Ours)                         & 34.0                 & 30.6                 & 27.5                 &    3.0                  \\
\multicolumn{5}{l}{\textit{\underline{Llama-3-8B-Inst + Syn. Data + DPO}}}                                                                            \\
TOA~(Ours)                         & 52.2                 & \textbf{55.8}        & \textbf{39.2}        &   3.0                  \\ \bottomrule
\end{tabular}%
}
\caption{
Comparison of synthetic data generated by each baseline and fine-tuning Llama-3-8B-Inst. Using 4 small models, outputs are generated with Ultrafeedback~\cite{DBLP:conf/icml/CuiY0YH0NXXL0024} prompts, sampling 160 responses per prompt via ArmoRM, retaining the best. Avg. FLOPs per prompt are reported. $^\dagger$ is taken from \citet{DBLP:journals/corr/abs-2405-14734}. 
Fine-tuning results for Qwen2-7B-Instruct are detailed in Table\,\ref{table:fine-tune-synthetic-align-qwen}.
}
\label{table:fine-tune-synthetic-align-llama}
\end{table}

In Figure\,\ref{fig:reward_layer_by_layer}, we analyze the rewards of the optimal child response node for each model node at any given layer. The results show that for the same model, as the depth increases, the reward of its optimal child response node tends to grow within a certain range. This indicates that the refinement process can effectively improve response. We also observe that the optimal responses often appear in deeper layers globally. We display some built tree examples in Figure\,\ref{fig:example_tree_alpaca_eval_a}, \ref{fig:example_tree_alpaca_eval_b}, \ref{fig:example_tree_alpaca_eval_c}, and \ref{fig:example_tree_alpaca_eval_d}.

\noindent{\textbf{Reward hacking exists in TOA.}} 
We use a learned reward model to guide TOA, optimizing output samples to the reward model but risking reward hacking. As shown in Figure\,\ref{fig:scaling_nmt}, the left figure illustrates a steady improvement in the reward of generated translations with increased compute. However, the right figure reveals that when re-evaluated by an external model, the score initially improves but later declines, indicating reward hacking. However, TOA remains the best overall approach. Similarly, \emph{PRS}, which also employs a reward model, encounters the same issue. These findings underscore more robust reward models.

\noindent{\textbf{TOA achieves competitive results on AlpacaEval and SOTA performance on WMT.}} \ With best-of-\( N \) sampling, TOA attains a 72.2\% LC Win Rate on AlpacaEval (Table\,\ref{tab:alpaca_eval}), surpassing all baselines. On WMT'21 and WMT'22 (Table\,\ref{table:part_nmt}), it sets a new SOTA benchmark, outperforming GPT-4 and CPO~\cite{xu2024contrastive}. While MoA ranks second, it demands much higher computational resources.

\noindent{\textbf{Enhanced data synthesis is enabled using TOA.}} We conduct practical tests to evaluate the effectiveness of synthetic data generated by various methods. In the alignment task, we use Ultrafeedback input prompts to produce synthetic outputs. 160 outputs are generated for each prompt, and the response with the highest reward, as determined by ArmoRM, is retained. The model is then fine-tuned using this generated data. As shown in Table\,\ref{table:fine-tune-synthetic-align-llama}, synthetic data generated by our method delivers the best results on AlpacaEval and Arena-Hard. 

We further train the model using DPO~\cite{DBLP:conf/nips/RafailovSMMEF23}, selecting the highest-reward response as the positive sample and the 30th-ranked one as the reject sample, based on optimal ranking experiments. Post-DPO training, our method sets a new SOTA, outperforming SimPO~\cite{DBLP:journals/corr/abs-2405-14734}. More results are included in Appendix~\ref{sec:further_analysis}.

\section{Conclusion}
We study scaling inference compute with multi-agent systems. We first formalize previous work on model combinations in a unified framework. We then propose TOA, a novel method to leverage MCTS for model coordination. Extensive experiments reveal multi-agent scaling laws, with our method achieving the best efficiency in synthetic data generation and scaling inference compute.

\section{Limitations}
Although multi-agent sampling is more compute-optimal than single-agent sampling, it requires a better design to reduce GPU memory usage when we need to load the model locally. Currently, we are using a straightforward approach, which involves loading different models simultaneously during the generation phase. This results in higher GPU memory consumption. However, if we directly call cloud-based API services, we do not need to host the models locally, and our method can significantly reduce the number of API calls while achieving the same level of performance.

\section*{Acknowledgements}
This research is supported by the National Research Foundation, Singapore under its AI Singapore Programme (AISG Award No: AISG2-PhD-2021-08-016[T]).

\bibliography{anthology,custom}

\clearpage
\appendix

\section{More Related Work}\label{sec:more_related_work}
\noindent{\textbf{Monte Carlo Tree Search.}} \ Monte Carlo Tree Search (MCTS)~\cite{DBLP:journals/tciaig/BrownePWLCRTPSC12} has been effectively utilized for advancing large language models. \citet{DBLP:conf/icml/WanFWM00024} leverage MCTS for token-level language model inference, while \citet{DBLP:journals/corr/abs-2404-12253} employ it to explore reasoning paths, facilitating self-improvement. In our work, we are the \emph{first} to apply MCTS for model coordination, introducing a novel tree structure that alternates between model layers and response layers.

\noindent{\textbf{Synthetic Data Generation.}} \ Generating data to advance AI systems during both pre-training and post-training stages is increasingly important.
\citet{DBLP:conf/icml/DohmatobFYCK24} highlight a challenging scaling law, showing that excessive pre-training on synthetic data leads to performance decay. \citet{DBLP:conf/naacl/GuoSVC24} observe that continual training on self-generated data reduces linguistic diversity. Thus, optimizing the use of synthetic data has become a critical focus. \citet{DBLP:journals/corr/abs-2404-01413} find that mixing synthetic data with real data enhances pre-training performance compared to using only real data. \citet{DBLP:journals/corr/abs-2406-07515} propose incorporating a verifier to prevent model collapse, while \citet{DBLP:journals/corr/abs-2406-20094} use personas to improve the diversity of generated synthetic data. Reject sampling or best-of-\(N\) sampling~\cite{DBLP:conf/iclr/0002ZJKSLL24} employs increased inference computation to generate more output samples, thereby raising the likelihood of obtaining strong samples. \citet{DBLP:conf/emnlp/YeN24} propose a preference-guided reflective mechanism to further improve best-of-\(N\) sampling.

\section{More Setup Details}

\subsection{Language Models}\label{sec:app_lang_model}
We combine models for evaluation. 
\noindent{\textbf{Small-scale combination}} has 4 models including: 
\begin{itemize}[itemsep=-5pt, topsep=0pt, leftmargin=0.3cm]
    \item \textbf{Llama-3.1-8B-Instruct}~\cite{llama3modelcard}: \\ \url{https://huggingface.co/meta-llama/Llama-3.1-8B-Instruct};

    \item \textbf{Qwen2-7B-Instruct}~\cite{qwen2}: \\ \url{https://huggingface.co/Qwen/Qwen2-7B-Instruct};

    \item \textbf{Mistral-7B-Instruct-v0.2}~\cite{jiang2023mistral}: \\ \url{https://huggingface.co/mistralai/Mistral-7B-Instruct-v0.2};

    \item \textbf{Yi-1.5-9B-Chat-16K}~\cite{DBLP:journals/corr/abs-2403-04652}: \\ \url{https://huggingface.co/01-ai/Yi-1.5-9B-Chat-16K}.
\end{itemize}

\noindent{\textbf{Larg-scale combination}} comprises 5 models including:
\begin{itemize}[itemsep=-5pt, topsep=0pt, leftmargin=0.3cm]
    \item \textbf{Llama-3.1-70B-Instruct}~\cite{llama3modelcard}: \\ \url{https://huggingface.co/meta-llama/Llama-3.1-70B-Instruct};

    \item \textbf{Mistral-Large-Instruct-2407}~\cite{mistral2024large}: \\ \url{https://huggingface.co/mistralai/Mistral-Large-Instruct-2407};

    \item \textbf{Qwen2-72B-Instruct}~\cite{qwen2}: \\ \url{https://huggingface.co/Qwen/Qwen2-72B-Instruct};

    \item \textbf{Mixtral-8x22B-Instruct-v0.1}~\cite{mistral2024mixtral}: \\ \url{https://huggingface.co/mistralai/Mixtral-8x22B-Instruct-v0.1};

    \item \textbf{Wizardlm-2-8x22b}~\cite{xu2024wizardlm}: \\ \url{https://huggingface.co/alpindale/WizardLM-2-8x22B}. 
\end{itemize}

\begin{table}[!t]
\centering
\resizebox{\linewidth}{!}{%
\begin{tabular}{lllc}
\hline
\textbf{Tasks}             & \textbf{Datasets}       &                             & \textbf{Size} \\ \hline
\multirow{5}{*}{NMT}       & WMT’21                  & \multicolumn{1}{c}{is - en} & 1000               \\ \cline{2-4} 
                           & \multirow{4}{*}{WMT’22} & \multicolumn{1}{c}{zh - en} & 1875               \\
                           &                         & \multicolumn{1}{c}{ru - en} & 2016               \\
                           &                         & \multicolumn{1}{c}{de - en} & 1984               \\
                           &                         & \multicolumn{1}{c}{cs - en} & 1448               \\ \hline
\multirow{2}{*}{Alignment} & AlpacaEval              &                             & 805                \\ 
                           & Arena-Hard              &                             & 500                \\ \hline
Reasoning                  & MATH                    &                             & 100                \\ \hline
Data Synthesis             & Ultrafeedback           &                             & 59876              \\ \hline
\end{tabular}%
}
\caption{Statistics of tasks and datasets used in our work. AlpacaEval is from \url{https://tatsu-lab.github.io/alpaca_eval/} and Arena-Hard is from \url{https://github.com/lmarena/arena-hard-auto}. We randomly sample 100 problems from MATH500~\cite{DBLP:conf/iclr/LightmanKBEBLLS24}. For Ultrafeedback, we use the version of \texttt{princeton-nlp/llama3-ultrafeedback-armorm} processed by \citet{xu2024contrastive}. }
\label{table:datasets}
\end{table}

\subsection{Statistics of Tasks and Datasets}\label{statistics}
Table\,\ref{table:datasets} provides details of the tasks and datasets used in this paper.

\subsection{Hyper-parameter Settings}\label{sec:setting_inference}
We present the hyper-parameters used across different tasks for our TOA:
\begin{itemize}[itemsep=-5pt, topsep=0pt, leftmargin=0.3cm]
    \item \textbf{Alignment:} The maximum layer width is set to \( \lfloor \frac{N}{3} \rfloor \), and \( \alpha \) is set to 0.01. 
    \item \textbf{Machine Translation:} The maximum layer width is also set to \( \lfloor \frac{N}{3} \rfloor \), with \( \alpha \) set to 0.05.  
    \item \textbf{Math Reasoning} The maximum layer width is set to \( \lfloor \frac{N}{2} \rfloor \), and \( \alpha \) is set to 0.01.
\end{itemize}

The temperature in sampling is set to 0.7, and \texttt{top\_p} is set to 1.
For MoA, we use the default prompt for generation when alignment is involved.

\subsection{Prompts Used for Generation}\label{sec:setting_prompt_refine}
\begin{figure}[!t]
\centering
\includegraphics[width=0.8\columnwidth]{ 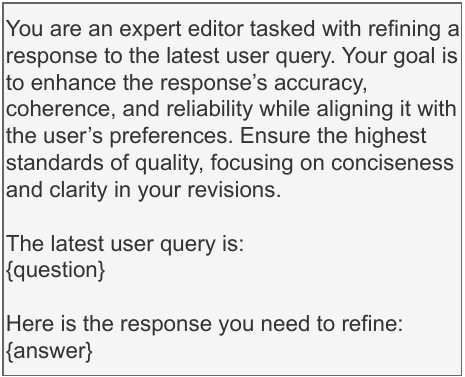} 
\caption{The refinement prompt for alignment.}
\label{fig:prompt_refine_alignment}
\end{figure}

\begin{figure}[!t]
\centering
\includegraphics[width=0.8\columnwidth]{ 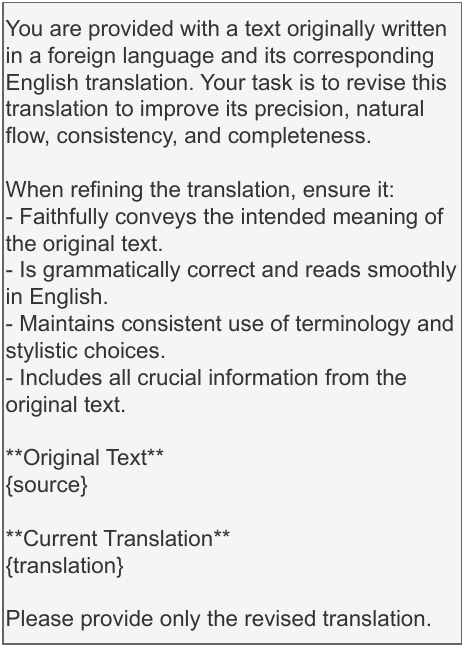} 
\caption{The refinement prompt for machine translation.}
\label{fig:prompt_refine_mt}
\end{figure}

\begin{figure}[!t]
\centering
\includegraphics[width=0.8\columnwidth]{ 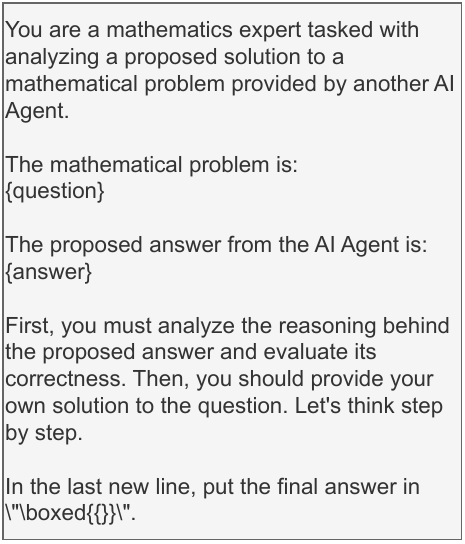} 
\caption{The refinement prompt for math reasoning.}
\label{fig:prompt_refine_math}
\end{figure}

Figures\,\ref{fig:prompt_refine_alignment}, \ref{fig:prompt_refine_mt}, and \ref{fig:prompt_refine_math} provide the refinement prompts that we use for the tasks of alignment, machine translation, and math reasoning, respectively.

\begin{table}[!t]
\centering
\resizebox{\linewidth}{!}{%
\begin{tabular}{lccc}
\toprule
                                                & \multicolumn{1}{l}{} & \multicolumn{2}{c}{\textbf{AlpacaEval 2.0}} \\
                  & \textbf{Size}       & \textbf{LC WR}         & \textbf{WR}        \\ \midrule
Llama-3-8B-Instruct                             & -                    & 24.8                   & 23.6               \\ \midrule
\multicolumn{4}{l}{\textit{\underline{Llama-3-8B-Base + Syn. Data + SFT}}}                                              \\
Magpie-Air-300K-Raw                                    & 300K                 & 22.0                   & 21.7               \\
Magpie-Air-300K-Filtered                               & 300K                 & 22.7                   & 24.0               \\
TOA (Ours)      & 80K                  & 24.7                   & 27.0               \\ \midrule
\multicolumn{4}{l}{\textit{\underline{Llama-3-8B-Inst + Syn. Data + SFT}}}                                              \\ 
TOA (Ours) & 80K                  & \textbf{32.8}                   & \textbf{33.3}              \\ \bottomrule
\end{tabular}%
}
\caption{Comparison with Magpie~\cite{DBLP:journals/corr/abs-2406-08464} for alignment data synthesis. We use the prompts from Magpie (\texttt{Magpie-Align/Magpie-Air-300K-Filtered}), and then use TOA (with 4 small-scale models) to generate synthetic outputs. We sample 80 outputs per prompt and use ArmoRM to keep the best one. We randomly sample 80K prompts from Magpie.}
\label{tab:compare_with_magpie}
\end{table}

\begin{figure}[!t]
\centering
\includegraphics[width=\linewidth]{ 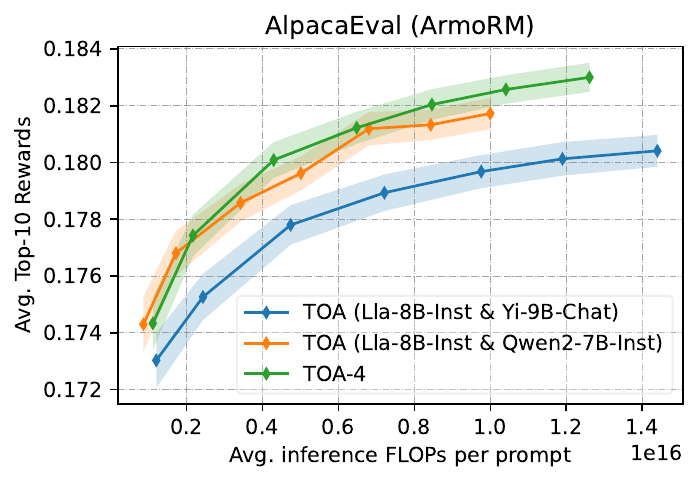}
\caption{Ablation of the number of models for combination in TOA. We use the small-scale sets for evaluation.
}
\label{fig:ablation_model_num}
\end{figure}

\section{Further Analysis}\label{sec:further_analysis}
\noindent{\textbf{Comparison with Magpie.}} \ 
To evaluate the performance of our method, we conduct a comparison with Magpie~\cite{DBLP:journals/corr/abs-2406-08464}, a recent approach for synthetic alignment data generation. For a fair comparison, we utilize the dataset provided by Magpie, specifically \texttt{Magpie-Align/Magpie-Air-300K-Filtered}. We randomly sample 80K prompts and employ a group of small-scale models to generate synthetic outputs, using ArmoRM-Llama3-8B-v0.1 as the reward model. Each prompt samples 80 outputs. For each prompt, only the highest-quality response is retained for training purposes. As shown in Table\,\ref{tab:compare_with_magpie}, our method consistently outperforms Magpie, showing our method's enhanced effectiveness and robustness.

\begin{figure}[!t]
\centering
\includegraphics[width=\linewidth]{ 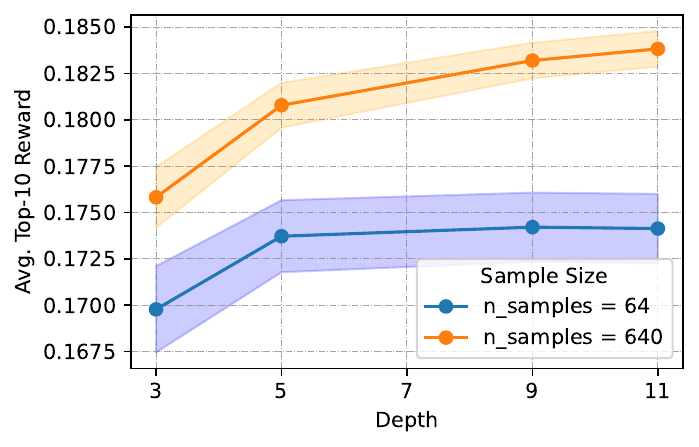}
\caption{Effect of depth for TOA. We vary the maximum depth of the tree between 3, 5, 9, and 11. The top 10 average reward (using ArmoRM) is reported on AlpacaEval with 200 samples.
}
\label{fig:effect_depth}
\end{figure}

\noindent{\textbf{Effect of Number of Models.}} \ 
We investigate the impact of the number of models used for combination. In Figure\,\ref{fig:ablation_model_num}, we reduce the number of models from 4 to 2 and evaluate in two different settings. Our results indicate that using 2 models does not provide greater computational efficiency compared to using 4 models.

\noindent{\textbf{Effect of Tree Depth for TOA.}} \ 
As illustrated in Figure\,\ref{fig:effect_depth}, we analyze the impact of tree depth on the performance of our TOA method. Our findings reveal that with a larger total number of samples for generation, such as 640, greater tree depth results in improved performance. Conversely, when the sample size is limited, such as 64, a smaller tree depth proves to be more effective. This behavior arises because TOA requires a broader layer width to ensure sufficient exploration.

\section{Supplementary Experiments}

\begin{figure}[!t]
\centering
\begin{minipage}{\linewidth}
    \centering
        \includegraphics[width=0.95\linewidth]{ 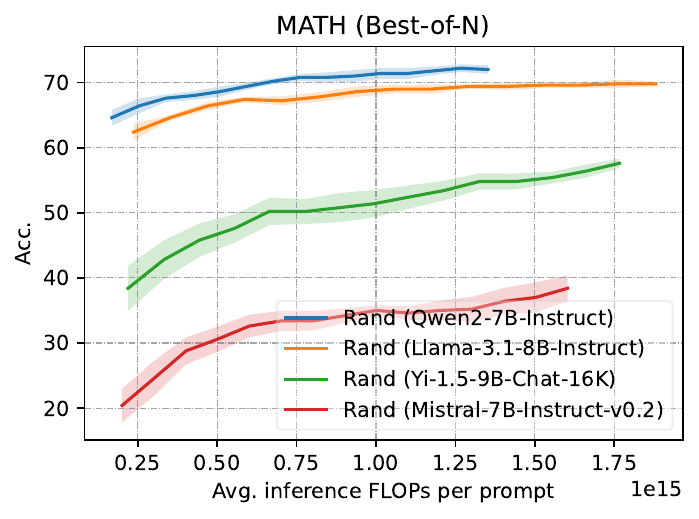}
\end{minipage}
\begin{minipage}{\linewidth}
    \centering
        \includegraphics[width=0.95\linewidth]{ 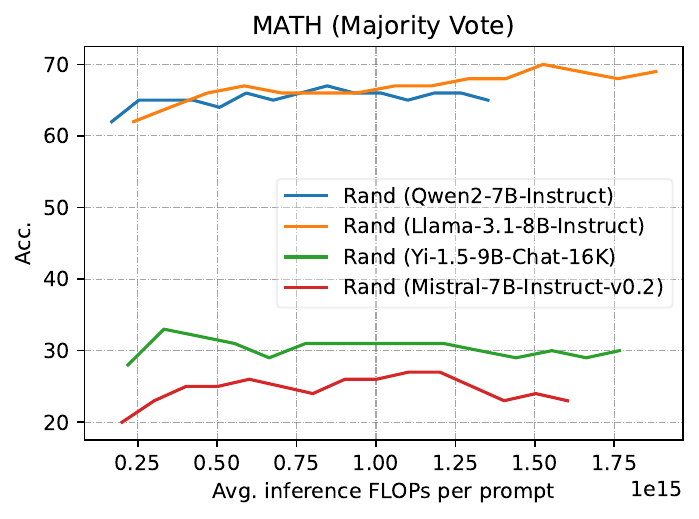}
\end{minipage}
\caption{
Performance of different language models on MATH.
}
\label{fig:math_result_differ_model}
\end{figure}

We present additional experiments to complement the primary experiments detailed in $\S$\,\ref{results} of the main paper. Below is a concise overview of the supplementary tables and figures, with detailed analyses already provided in $\S$\,\ref{results}.

\begin{figure}
\centering
\begin{minipage}{\linewidth}
    \centering
        \includegraphics[width=\linewidth]{ 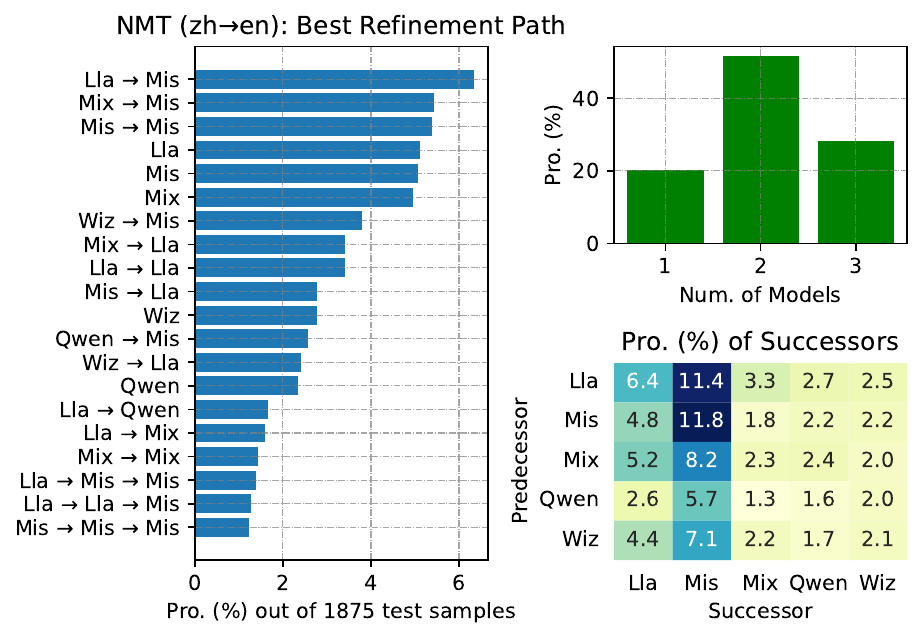}
\end{minipage}
\caption{On the machine translation task, for TOA with 5 large models, we identify the best refinement paths per prompt, maximizing reward. We present the top 20 most frequent paths, the number of models used for generation, and the successor proportions for each predecessor.}
\label{fig:refine_path_mt}
\end{figure}

\begin{figure}[t]
\centering
\includegraphics[width=23cm, height=7cm, angle=90]{ 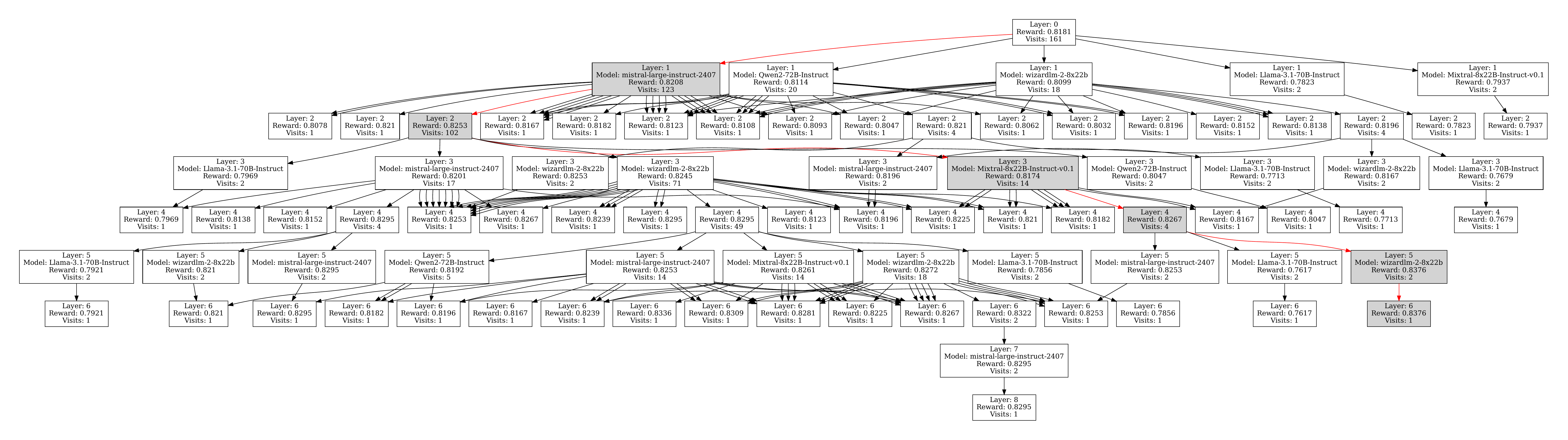}
\caption{Example (a) of a decision tree constructed using TOA. Best viewed with zooming in.}
\label{fig:example_tree_alpaca_eval_a}
\end{figure}

\begin{figure}[t]
\centering
\includegraphics[width=23cm, height=4cm, angle=90]{ 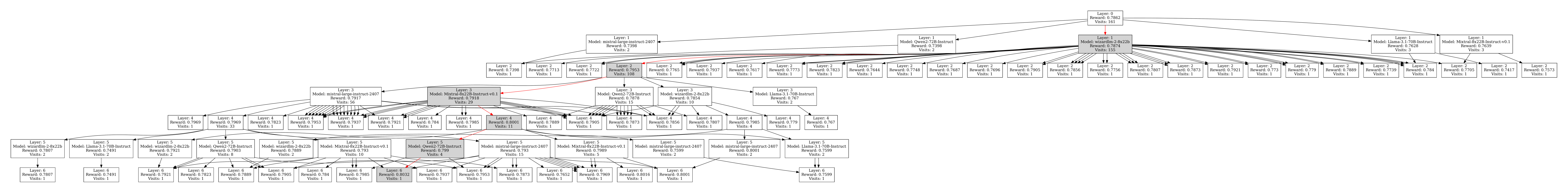}
\caption{Example (b) of a decision tree constructed using TOA. Best viewed with zooming in.}
\label{fig:example_tree_alpaca_eval_b}
\end{figure}

\begin{figure}[t]
\centering
\includegraphics[width=23cm, height=5cm, angle=90]{ 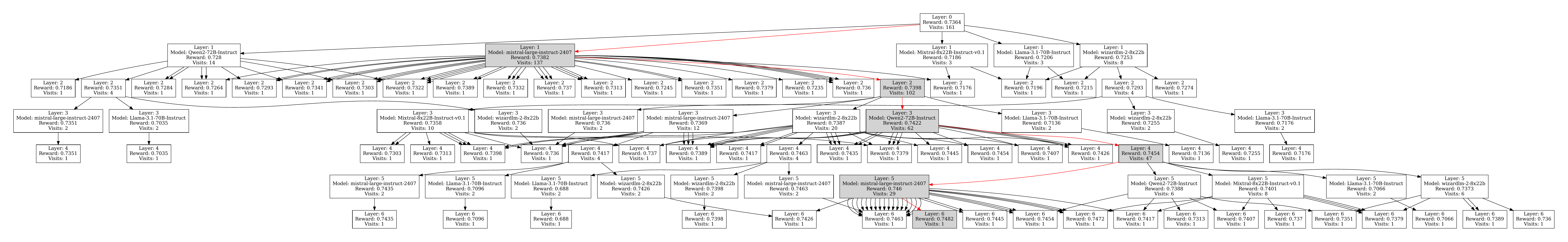}
\caption{Example (c) of a decision tree constructed using TOA. Best viewed with zooming in.}
\label{fig:example_tree_alpaca_eval_c}
\end{figure}

\begin{figure}[t]
\centering
\includegraphics[width=23cm, height=5cm, angle=90]{ 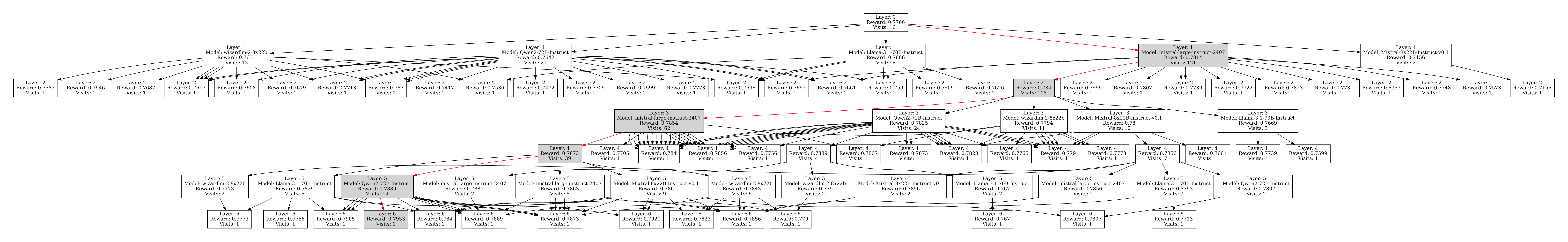}
\caption{Example (d) of a decision tree constructed using TOA. Best viewed with zooming in.}
\label{fig:example_tree_alpaca_eval_d}
\end{figure}

Figure\,\ref{fig:math_result_differ_model} presents the evaluation results on the MATH test set for different language models.
Figure\,\ref{fig:refine_path_mt} is the counterpart of Figure\,\ref{fig:refine_path_alpaca_eval} for machine translation, to dissect the optimal refinement paths per prompt.
Table\,\ref{table:full_nmt} demonstrates our state-of-the-art performance on the WMT benchmark.
Table\,\ref{table:fine-tune-synthetic-align-qwen}, in conjunction with Table\,\ref{table:fine-tune-synthetic-align-llama}, highlights our capability to enhance data synthesis.
Lastly, Figure\,\ref{fig:example_tree_alpaca_eval_a}, \ref{fig:example_tree_alpaca_eval_b}, \ref{fig:example_tree_alpaca_eval_c}, and \ref{fig:example_tree_alpaca_eval_d} show the construction of decision trees utilizing our methodology. The evaluation was conducted on the AlpacaEval dataset with five large-scale models. Each figure corresponds to a different question. The path leading to the highest reward (ArmoRM) is highlighted in red.

\begin{table*}[!t]
\centering
\Large
\resizebox{\textwidth}{!}{%
\begin{tabular}{lcccccccccccc}
\toprule
\multirow{2}{*}{Methods} & \multicolumn{2}{c}{\texttt{de}}         & \multicolumn{2}{c}{\texttt{cs}}         & \multicolumn{2}{c}{\texttt{is}}        & \multicolumn{2}{c}{\texttt{zh}}         & \multicolumn{2}{c}{\texttt{ru}}         & \multicolumn{2}{c}{\texttt{Avg.}} \\
                         & KXXL           & MetricX       & KXXL           & MetricX       & KXXL          & MetricX       & KXXL           & MetricX       & KXXL           & MetricX       & KXXL             & MetricX        \\ \midrule
Gold Reference           & 78.56          & -             & 83.11          & -             & 85.04         & -             & 74.19          & -             & 79.59          & -             & 80.10            & -              \\
WMT Winners              & 83.59          & 1.60          & 82.53          & 1.44          & 85.6          & 1.54          & 73.28          & 3.68          & 80.97          & 1.42          & 81.19            & 1.94           \\
GPT-4                    & 84.58          & 1.37          & 83.55          & 1.46          & \textbf{85.9} & 1.39          & 77.65          & 2.22          & 81.34          & 1.26          & 82.60            & 1.54           \\
CPO                      & 83.97          & 1.45          & 83.75          & 1.43          & 85.73         & \textbf{1.35} & 77.17          & 2.53          & 81.54          & 1.34          & 82.43            & 1.62           \\ \midrule
Rand (lla-3.1-70b)     & 85.41          & 1.32          & 85.09          & 1.33          & 84.45         & 1.76          & 79.07          & 2.23          & 82.21          & 1.20          & 83.24            & 1.57           \\
Rand (Mis-large-2407)  & 85.60          & 1.26          & 84.54          & 1.48          & 83.35         & 1.98          & 79.50          & 1.95          & 82.64          & 1.09          & 83.13            & 1.55           \\
\emph{PRS} (lla-3.1-70b)        & 85.83          & 1.18          & 85.71          & 1.18          & 84.76         & 1.51          & 79.51          & 1.86          & 82.67          & 1.04          & 83.70            & 1.36           \\
\emph{PRS} (Mis-large-2407)     & 86.18          & 1.17          & 85.72          & 1.33          & 84.01         & 1.74          & 79.82          & 1.79          & 82.68          & 1.03          & 83.68            & 1.41           \\
Par. Ensemble                 & 85.82          & 1.26          & 85.35          & 1.34          & 83.58         & 1.91          & 79.47          & 2.03          & 82.53          & 1.13          & 83.35            & 1.53           \\
Seq. Refine                      & 86.14          & 1.14          & 85.81          & 1.19          & 83.86         & 1.63          & 79.39          & 1.79          & 82.68          & \textbf{0.99} & 83.58            & 1.35           \\
MoA                      & 86.20          & 1.14          & 86.13          & 1.19          & 84.87         & 1.60          & 79.87          & 1.79          & 82.80          & 1.00          & 83.97            & 1.34           \\
TOA (Ours)                    & \textbf{86.25} & \textbf{1.11} & \textbf{86.31} & \textbf{1.16} & 84.81         & 1.50          & \textbf{79.89} & \textbf{1.77} & \textbf{83.00} & 1.00          & \textbf{84.05}   & \textbf{1.31}  \\ \bottomrule
\end{tabular}%
}
\caption{Results of xx-to-en translation on WMT21 and WMT22. \texttt{is}-to-\texttt{en} is from WMT21 and the rest is from WMT22. The results of Gold Reference, WMT Winners, GPT-4, and CPO are taken from \cite{xu2024contrastive}. In our experiments, to scale inference compute, for each test sample, we sample 160 translations and use KIWI~(\texttt{Unbabel/wmt22-cometkiwi-da}) to select the best one with the highest score. Then the selected samples will be re-evaluated by KIWI-XXL~(or KXXL)~(\texttt{Unbabel/wmt23-cometkiwi-da-xxl}) and MetricX~\cite{juraska-etal-2023-metricx}~(\texttt{google/metricx-23-xxl-v2p0}). For MetricX, the lower the better. Here, KIWI-22 and KIWI-XXL are reference-free evaluations and MetricX is a reference-based evaluation.}
\label{table:full_nmt}
\end{table*}

\begin{table}[!t]
\centering
\resizebox{\linewidth}{!}{%
\begin{tabular}{lcc}
\toprule
                                          & \multicolumn{2}{c}{\textbf{AlpacaEval}}     \\
                                          & \textbf{LC WR}       & \textbf{WR}          \\ \midrule
Qwen2-7B-Instruct                         & 21.2                 & 19.5                 \\ \midrule
\textit{\underline{{Qwen2-7B-Inst + Syn. Data + SFT}}   } & \multicolumn{1}{l}{} & \multicolumn{1}{l}{} \\
Rand (Qwen2-7B-Inst)                  & 25.0                 & 23.4                 \\
\emph{PRS} (Qwen2-7B-Inst)                   & 27.1                 & 22.6                 \\
Par. Ensemble                             & 20.2                 & 19.4                 \\
Seq. Refine                               & 25.9 &  22.9 \\
MoA                                       & 22.7                 & 24.1                 \\
TOA (Ours)                                      & 28.3                 & 25.6                 \\ 
\textit{\underline{{Qwen2-7B-Inst + Syn. Data + DPO}}} & \multicolumn{1}{l}{} & \multicolumn{1}{l}{} \\
TOA (Ours)                             &\textbf{34.4}                 & \textbf{50.5}  \\ \bottomrule            
\end{tabular}%
}
\caption{Qwen2-7B-Instruct fine-tuning results using different synthetic data.}
\label{table:fine-tune-synthetic-align-qwen}
\end{table}

\end{document}